\documentclass[11pt]{article}

\usepackage[final]{acl}

\usepackage{times}
\usepackage{latexsym}

\usepackage[T1]{fontenc}

\usepackage[utf8]{inputenc}

\usepackage{microtype}

\usepackage{inconsolata}

\usepackage{graphicx}

\usepackage{hyperref}
\usepackage{graphicx}
\usepackage{tcolorbox}
\usepackage{url}
\usepackage{xspace}
\usepackage{enumitem}
\usepackage{booktabs}
\usepackage{makecell}
\usepackage{pifont}
\usepackage{wrapfig}
\usepackage{caption}
\usepackage{algorithm}
\usepackage{algorithmic}
\usepackage[table]{xcolor}
\usepackage{multirow}
\usepackage{amsmath}
\usepackage{newfloat}
\usepackage{float}
\usepackage{listings}
\usepackage{xcolor}  

\lstdefinestyle{python_highlight}{
    language=Python,
    basicstyle=\ttfamily\small,       
    keywordstyle=\color{blue}\bfseries,  
    commentstyle=\color{gray}\itshape,   
    stringstyle=\color{orange},          
    breaklines=true,                     
    showstringspaces=false,              
    numbers=left,                        
    numberstyle=\tiny\color{gray},       
    frame=single,                        
    rulecolor=\color{black},             
    tabsize=4                            
}

\def\eg{\emph{e.g.,}\xspace}

\def\method{\textbf{EvoCoT}\xspace}

\title{\method: Overcoming the Exploration Bottleneck in Reinforcement Learning for LLMs}

\author{
Huanyu Liu$^{1}$, Jia Li$^{2}$, Yihong Dong$^{1}$, Chang Yu$^{1}$, Taozhi Chen$^{3}$, \\  \textbf{Lecheng Wang$^{1}$}, \textbf{Yongding Tao$^{1}$}, \textbf{Bin Gu$^{4}$}\thanks{Corresponding author.}, \textbf{Ge Li$^{1}$}\footnotemark[1]\\
$^1$ School of Computer Science, Peking University \quad $^2$ College of AI, Tsinghua University \\
$^3$ Imperial College London \quad $^4$ Institute of Software, Chinese Academy of Sciences \\
\texttt{huanyuliu@stu.pku.edu.cn} \quad \texttt{lige@pku.edu.cn}
}

\begin{document}
\maketitle

\begin{abstract}
Reinforcement learning with verifiable reward (RLVR) has become a promising paradigm for post-training large language models (LLMs) to improve their reasoning capability. However, when the rollout accuracy is low on hard problems, the reward becomes sparse, limiting learning efficiency and causing exploration bottlenecks. Existing approaches either rely on teacher models for distillation or filter out difficult problems, limiting scalability or restricting reasoning improvement through exploration.

We propose \method, a self-\textbf{\underline{Evo}}lving curriculum learning framework based on two-stage \textbf{\underline{C}}hain-\textbf{\underline{o}}f-\textbf{\underline{T}}hought (CoT) reasoning optimization. \method constrains the exploration space by self-generating and verifying CoT trajectories, then gradually shortens CoT steps to expand the space in a controlled way. The framework enables LLMs to stably learn from initially unsolved hard problems under sparse rewards. We apply \method to multiple LLM families, including Qwen, DeepSeek, and Llama. Experiments show that \method enables LLMs to solve previously unsolved problems, improves reasoning capability without external CoT supervision, and is compatible with various RL fine-tuning methods. We release the source code and models to support future at \url{https://github.com/gtxygyzb/EvoCoT}.

\end{abstract}

\section{Introduction}
\label{sec:introduction}

Recently, reinforcement learning with verifiable reward (RLVR) has emerged as a promising paradigm for the post-training of large language models (LLMs). LLMs demonstrate remarkable reasoning capability in solving complex tasks, from math problems to code generation. Existing works~\cite{deepseek_r1,ProRL} compute rewards via rule-based verification of predicted final answers, effectively enhancing reasoning capability without relying on annotated reasoning trajectories.

Within RLVR, we expect LLMs to explore correct reasoning trajectories during rollouts to obtain rewards and gradually improve their reasoning capability~\cite{DAPO,GRPO}. However, when the rollout accuracy is low on some hard problems, the LLM receives sparse rewards, hindering the improvement of reasoning capability. Due to the vast solution space, LLMs often face exploration bottlenecks on such problems.

In experiments, we find that LLMs often struggle to fully learn from hard problems, despite RLVR training. For example, even after sufficient training on the GSM8K~\cite{gsm8k} and MATH~\cite{math} training sets, Qwen2.5-7B still fails to solve 8.8\% and 22.0\% of the problems, respectively (see Table~\ref{tab:rq1_training_set}). These unsolved problems are still valuable for RLVR. If LLMs could exploit such problems more effectively during training, their reasoning capability could be further improved~\cite{ProRL}. 

Several recent works attempt to address this question. \ding{182}~One category of methods depends on teacher LLMs to provide hints or reasoning trajectories for \underline{\textbf{distillation}}~\cite{GuideGRPO, learning_RL_cant, LUFFY, TAPO, SRFT}. For instance, LUFFY~\cite{LUFFY} mixes outputs from teacher LLMs into the GRPO candidate set and applies importance sampling to emphasize low-probability but correct actions. These methods enhance performance but require access to teacher LLMs, which is a strong assumption that imposes high costs and limits scalability, especially when training flagship models without available teacher models. \ding{183}~Another category of methods attempts to control problem difficulty to facilitate curriculum learning for LLMs~\cite{Self-Evolving-Curriculum, RORL, AdaRFT}. RORL~\cite{RORL} computes the rollout accuracy for each group in a batch and retains only the problems within a predefined accuracy range. While this mitigates reward sparsity, it also \underline{\textbf{filters out}} many hard problems that could serve as valuable training data, restricting the LLM’s reasoning improvement through exploration. A detailed comparison is provided in Table~\ref{tab:RL_comparison}.

\begin{table}[t]
    \centering
    \caption{The comparison between existing reinforcement learning (RL) methods and \method.}
    \setlength{\tabcolsep}{2pt}
    \resizebox{1.0\linewidth}{!}{
    \begin{tabular}{lcc}
        \toprule
        Methods & \ding{182} Distillation-Free & \ding{183} Unfiltered \\
        \midrule
        ReLIFT \cite{learning_RL_cant} & \ding{55} & \ding{55}  \\
        AdaRFT \cite{AdaRFT} & \ding{51} & \ding{55}  \\
        RORL \cite{RORL} & \ding{51} & \ding{55}  \\
        TAPO \cite{TAPO} & \ding{55} & \ding{51}  \\
        LUFFY \cite{LUFFY} & \ding{55} & \ding{51}  \\
        Guide-GRPO \cite{GuideGRPO} & \ding{55} & \ding{55}  \\
        SRFT \cite{SRFT} & \ding{55} & \ding{51}  \\
        \midrule
        \method (Ours) & \ding{51} & \ding{51} \\
        \bottomrule
    \end{tabular}}
    \label{tab:RL_comparison}
\end{table}

In this paper, we aim to investigate the following question:

\begin{tcolorbox}[colback=gray!10!white, colframe=black, title=Key Question]
Can LLMs become self-evolving by overcoming exploration bottlenecks and progressively enhancing reasoning capability, without distillation from teacher models?

\end{tcolorbox}

The low rollout accuracy on hard problems is primarily due to the vast solution space, which substantially exceeds the current reasoning capability of LLMs~\cite{cot_without_prompt, train_long}, as shown in Figure~\ref{fig:framework}. We propose \method, a self-\textbf{\underline{Evo}}lving curriculum learning framework based on two-stage \textbf{\underline{C}}hain-\textbf{\underline{o}}f-\textbf{\underline{T}}hought (CoT) reasoning optimization. The core idea of \method is to constrain the size of the exploration space. In Stage 1, the LLM receives problems and final answers, and generates its own CoT trajectories. These CoTs are filtered and verified to construct step-by-step reasoning. In Stage 2, \method performs curriculum learning by progressively removing reasoning steps from each CoT trajectory. This step-wise reduction gradually expands the exploration space in a controlled manner, increasing reasoning difficulty while enabling stable training under sparse rewards. Through self-evolving iterations, the LLM enhances its reasoning capability and generates higher-quality CoTs, progressively solving a portion of initially unsolved hard problems.

We apply \method to LLMs across diverse model families, including Qwen, DeepSeek, Llama, and DeepSeek-R1-Distill-Qwen (referred to as R1-Qwen). Experimental results demonstrate that:

\begin{itemize}[leftmargin=*]
    \item Compared to GRPO, \method enables LLMs to overcome exploration bottlenecks on previously unsolved training set problems, with average improvements of +4.5 for Qwen2.5-7B and +21.7 for R1-Qwen-1.5B.
    \item Beyond the training set, \method transfers its learned reasoning to other math benchmarks, outperforming SimpleRL with average improvements of +2.3 on Qwen2.5-7B and +2.1 on R1-Qwen-1.5B.
    \item Compared to SFT and GRPO, \method supports more effective self-exploration, achieving average improvements of +10.8 and +1.6 across all evaluated LLMs.
\end{itemize}

\section{Related Work}
\label{sec:related_work}

\subsection{Reinforcement Learning with Verifiable Reward}

RLVR for LLMs has drawn considerable research attention following DeepSeek-R1~\cite{deepseek_r1} and Kimi-k1.5~\cite{kimi_k1.5}. However, recent studies~\cite{limit_of_rlvr,Echo_Chamber} suggest that the performance of the RLVR-trained model is fundamentally constrained by the base model's inherent capability, as RLVR only biases the base model's output distribution toward reward-maximizing paths. 
In RLVR, rewards are sometimes too sparse compared to the large solution space, causing exploration bottlenecks that prevent finding solutions unexplored by the base model.
Some works~\cite{learning_RL_cant,SASR,SRFT,SuperRL,TAPO,LUFFY,GuideGRPO,SWiRL} attempt to incorporate off-policy data into training. For instance, ReLIFT~\cite{learning_RL_cant}, SASR~\cite{SASR}, SRFT~\cite{SRFT} and SuperRL~\cite{SuperRL} integrate RLVR with supervised fine-tuning (SFT). Meanwhile, TAPO~\cite{TAPO}, LUFFY~\cite{LUFFY} and Guide-GRPO~\cite{GuideGRPO} leverage reference CoT or hints generated by teacher models, or query an external thought library to guide policy optimization. Unfortunately, these methods either rely on distillation from teacher models or high-quality training data.

\subsection{Curriculum Learning for Reasoning Tasks}

Curriculum learning~\cite{Curriculum_learning} is a training strategy that arranges examples ordered from easy to hard.
In RL, curriculum learning explores strategies to balance exploration and exploitation, with methods such as promising initialization~\cite{CL_1} and reverse curriculum generation~\cite{CL_2} showing effectiveness. However, in LLMs, overcoming exploration bottlenecks remains a major question.
Previous works~\cite{kimi_k1.5,Logic-RL,SATURN} explore the application of curriculum learning in RLVR for LLM post-training, demonstrating that the difficulty arrangement of the RL training data is critical for achieving competitive performance. However, existing difficulty-arranging methods have some limitations. RORL~\cite{RORL} filters out too hard or too easy problems for the current LLM to solve, but some discarded hard problems could be valuable for training;  E2H~\cite{E2H}, SEC~\cite{Self-Evolving-Curriculum} and AdaRFT~\cite{AdaRFT} dynamically adapt the probability distribution on difficulties for sampling, but they require fine-grained difficulty estimation in the dataset; R3~\cite{reverse_RL} and AdaBack~\cite{AdaBack} smoothly increase difficulty by showing the LLM gradually shorter prefixes of CoT, whereas they necessitate complete CoT data for training.



\section{EvoCoT}
\label{sec:EvoCoT}
\subsection{Self-Evolving Curriculum Learning Framework}
\label{sec:3_1}

\begin{figure*}[t]
\centering
\includegraphics[width=1\textwidth]{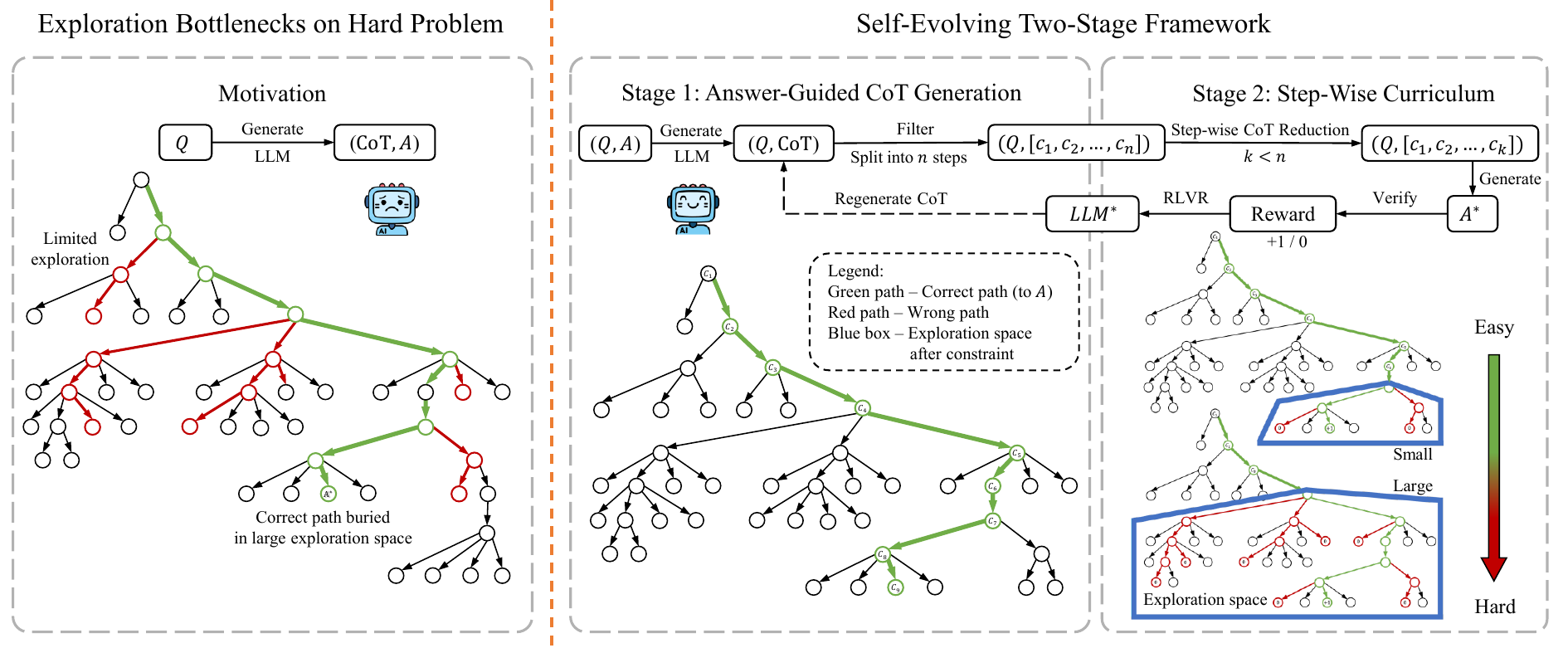}
\caption{The overall framework of \method. It is structured as two nested stages: 
\textbf{Stage 1: Answer-Guided Reasoning Path Self-Generation}, which generates and filters CoT trajectories from final-answer supervision, and 
\textbf{Stage 2: Step-Wise Curriculum Learning}, which implements curriculum learning by progressively shortening CoTs to increase difficulty and exploration space. The two stages iterate jointly, enabling the LLM to gradually enhance its reasoning capability through self-evolving optimization.}
\label{fig:framework}
\end{figure*}

We introduce \method, a self-evolving curriculum learning framework for LLMs. \method improves LLMs' reasoning capability through iterative training with gradually increasing difficulty. The core idea of \method is to constrain and gradually expand the exploration space. As illustrated in Figure~\ref{fig:framework}, \method is structured as two nested stages: \textbf{Stage 1: Answer-Guided Reasoning Path Self-Generation} constructs CoT trajectories from final answers, and \textbf{Stage 2: Step-Wise Curriculum Learning} implements step-wise CoT reduction for RLVR. The two stages iterate jointly, forming a self-evolving framework. And the overall pseudocode is provided in Appendix~\ref{app:pseudocode}. 



The following subsections respectively introduce: \ding{182} how CoTs are generated and filtered in Stage 1; \ding{183} how curriculum learning is implemented in Stage 2 via step-wise CoT reduction; and \ding{184} the self-evolving iterative optimization along with the advantages of \method.

\subsection{Stage 1: Answer-Guided Reasoning Path Self-Generation}
\label{sec:3_2}

Given a training dataset consisting of questions and final answers, the LLM generates CoT trajectories that reconstruct how the answer could be derived. This stage follows the intuition that reasoning paths are easier to construct when the final answer is provided. The generated CoTs are filtered to ensure logical consistency and are organized into multi-step trajectories connecting the question to the final answer. Importantly, this stage does not require annotated CoTs or teacher models, and transforms outcome-supervised data into reasoning paths in a fully self-generated manner.


The training input consists of math problems formatted as $(Q, A)$ pairs, where $Q$ is the question and $A$ is the final answer. No CoT annotations or distilled data are required. For each $(Q, A)$, the LLM is prompted to generate a reasoning chain $\hat{C}$ (detailed in Appendix~\ref{app:prompt_template}):

\begin{equation}
\footnotesize
(Q, A) \xrightarrow[\text{LLM}]{} \hat{C}
\end{equation}

Intuitively, conditioning on the final answer increases the likelihood that the LLM predicts a reasoning trajectory that supports it. To verify consistency, we check whether the LLM can derive the correct answer $A$ when conditioned on $(Q, \hat{C})$:

\begin{equation}
\footnotesize
(Q, \hat{C}) \xrightarrow[\text{LLM}]{} \hat{A}, \quad \text{retain } \hat{C} \text{ if } \hat{A} = A.
\end{equation}

Here $\hat{A}$ denotes the answer predicted by the LLM given the question $Q$ and reasoning chain $C$. Only reasoning chains that derive the correct answer are retained. Each verified $C$ is then split into step-wise format using the delimiter \texttt{"\textbackslash n\textbackslash n"}:

\begin{equation}
\footnotesize
(Q, \hat{C}) \xrightarrow[\text{split}]{} (Q, \hat{C} = \{\hat{c_1}, \hat{c_2}, ..., \hat{c_n}\})
\end{equation}

where each $c_i$ is a clear reasoning sub-step, forming a multi-step trajectory suitable for subsequent curriculum learning. \textbf{No additional constraints are applied to the self-generated $\hat{C}$, allowing the LLM to explore freely.}


\subsection{Stage 2: Step-Wise Curriculum Learning}
\label{sec:3_3}


Given the reasoning paths constructed in Stage 1, Stage 2 implements the curriculum learning by progressively shortening each CoT trajectory. Starting from complete CoT trajectories, \method gradually removes reasoning steps in reverse order, producing a series of training samples with increasing difficulty. As shown in Figure~\ref{fig:framework}, shorter CoTs expand the LLM's exploration space, making the reasoning more challenging. The step-wise reduction forms a difficulty progression, from easy samples with full guidance to hard ones requiring more exploration. Each sample is then used for rollouts to fine-tune the LLM with RLVR. 


Given a complete reasoning trajectory $(Q, c_1, c_2, \dots, c_n)$, training proceeds by gradually truncating the tail steps to increase difficulty. During each step-wise rollout, the curriculum follows:

\begin{equation}
\begin{footnotesize}
\begin{aligned}
&(Q, \hat{c_1}, \dots, \hat{c_{n}}) \xrightarrow[\text{roll out}]{} (\tilde{C}, \tilde{A}) \\
&(Q, \hat{c_1}, \dots, \hat{c_{n-1}}) \xrightarrow[\text{roll out}]{} (\tilde{C}, \tilde{A}) \\
&\quad\vdots \\
&(Q, \hat{c_1}) \xrightarrow[\text{roll out}]{} (\tilde{C}, \tilde{A}) \\
&(Q) \xrightarrow[\text{roll out}]{} (\tilde{C}, \tilde{A})
\end{aligned}
\end{footnotesize}
\end{equation}

where $(\tilde{C}, \tilde{A})$ denotes the CoT and answer rolled out by the LLM for RL training. Rollouts follow the provided partial CoT as prefix but the remaining steps 
are unconstrained. Starting from full-length CoTs, the LLM learns to generate correct answers under strong guidance. Gradually removing steps expands the exploration space of the LLM, increasing difficulty and encouraging the discovery of more complex reasoning paths. The step-wise curriculum within each sample stabilizes training under sparse rewards and improves the overall reasoning capability of the LLM.

Our design is motivated by two considerations:
\ding{182} Training with longer CoT guidance is easier than with shorter or no CoT, making the progressive reduction of steps a natural curriculum.
\ding{183} As trajectories shorten, the LLM needs to complement reasoning steps and ultimately derive $A$ directly from $Q$, which avoids reward hacking caused by revealing answers in the self-generated CoTs.

\subsection{Self-Evolving Iterative Optimization}
\label{sec:3_4}

\method follows a self-evolving two-stage process. In each iteration, the current LLM first generates CoT trajectories from $(Q, A)$ pairs (Stage 1). These CoTs are filtered and split into step-wise reasoning paths. Then, the LLM is trained via curriculum learning by progressively shortening the CoTs (Stage 2), increasing task difficulty. After updating the LLM’s parameters, its reasoning capability improves, enabling the generation of higher-quality CoTs in the next iteration. We use $\mathcal{Q}$, $\mathcal{A}$, and $\mathcal{\hat{C}}$ to denote the complete datasets. The $t$-th iteration can be represented as:

\begin{equation}
\begin{footnotesize}
\begin{aligned}
\footnotesize
\mathcal{\hat{C}}^{(t)} &= \mathrm{Generate}\!\left(\mathcal{Q}, \mathcal{A};\, \mathrm{LLM}^{(t)}\right), \\
\mathrm{LLM}^{(t+1)} &= \mathrm{Train}\!\left(\mathcal{Q}, \mathcal{\hat{C}}^{(t)} \mathcal{A}; \mathrm{LLM}^{(t)} \right) \\
\end{aligned}
\end{footnotesize}
\end{equation}

Although initial CoTs may be imperfect, iterative training can improve the LLM's reasoning capability and lead to better CoT generation, which in turn provides stronger guidance for subsequent learning. Over multiple iterations, our self-evolving \method enhances both the quality of generated reasoning and the LLM’s overall reasoning capability.

\textbf{Note that} \method is orthogonal to existing training paradigms, such as DAPO~\cite{DAPO}, and can be applied as a complementary stage after post-training. This orthogonality arises from its self-exploration process, which does not rely on external supervision. Rather than replacing prior methods like GRPO, \method further enhances reasoning through iterative self-evolution. Further experimental results with DAPO are provided in Appendix~\ref{app:dapo}.

\method has three main advantages:

\begin{itemize}[leftmargin=*]
    \item \textbf{Avoiding reliance on human-annotated CoTs:} The LLM learns solely from automatically generated reasoning chains based on $(Q, A)$ pairs, without requiring any manual CoT labels or teacher models.

    \item \textbf{Reducing the risk of failure on hard problems with large exploration space:} Step-wise CoT reduction gradually increases the difficulty by expanding the LLM's exploration space, enabling more stable learning under sparse rewards.
    
    \item \textbf{Eliminating the need to manually build training data ordered by difficulty:} Each single CoT sample naturally supports curriculum learning.

\end{itemize}

\section{Experiments}
\label{sec:experiments}

We conduct a large-scale experiment to evaluate \method. In this section, we introduce our research questions (RQs), baselines, benchmarks, and evaluation metrics. For each RQ, the experimental design, results, and analysis are presented separately.

\subsection{Research Questions}
\label{sec:experiment:RQs}

Our experimental study is guided by the following research questions:

\textbf{RQ1: Can \method solve previously unsolved training problems?}  
We evaluate whether \method enables LLMs to correctly solve problems in the training set that were initially unsolved, verifying its effectiveness in overcoming exploration bottlenecks.

\textbf{RQ2: Can \method enhance LLMs' reasoning on unseen math problems?}  
We evaluate whether \method enhances the LLM's performance on a diverse set of math benchmarks that are not included in the training data.

\textbf{RQ3: How effective is \method compared to other learning paradigms?}
We compare \method with RLVR and supervised fine-tuning (SFT) to isolate the effectiveness of the self-exploration in \method.

\textbf{RQ4: Can \method indefinitely improve reasoning through self-evolution?}  
We evaluate whether \method can continuously enhance LLM reasoning through iteration, or if the performance saturates, revealing its scalability and inherent limitations.

\subsection{Experimental Setup}
\label{sec:setup}

\paragraph{Baselines.} 
We compare \method with recent open-source RLVR works, including SimpleRL~\cite{SimpleRL}, DeepScaleR~\cite{deepscaler2025}, and Open-Reasoner-Zero~\cite{Open-Reasoner-Zero}. In addition to vanilla GRPO, we include PRIME~\cite{prime}, AdaRFT~\cite{AdaRFT}, and SEC~\cite{Self-Evolving-Curriculum} as method-level baselines, representing recent RL or curriculum-learning improvements without distillation. For fairness, we use the released LLMs with the prompt templates reported in the original papers, and all LLMs share the same sampling settings. We also ensure that SFT and GRPO use the same number of training steps.

\paragraph{\method Hyperparameters}

We apply \method across diverse model families, including Qwen2.5-7B \cite{qwen2.5}, Llama3.1-8B \cite{llama3}, DeepSeek-Math-7B \cite{GRPO}, and DeepSeek-R1-Distill-Qwen-1.5B (referred to as R1-Qwen-1.5B) \cite{deepseek_r1}. We follow the baseline models and training setup provided by DeepScaleR and SimpleRL-Zoo\footnote{\url{https://github.com/volcengine/verl}}.
\ding{182} In Stage 1, we collect problems from the GSM8K and MATH training sets where the LLM fails to solve the problem in all 8 rollouts. For each unsolved problem, 8 reasoning paths are sampled with a temperature of 1.0.
\ding{183} In Stage 2, detailed training hyperparameters are provided in Appendix~\ref{app:training_eval}. Since the number of failed problems varies across LLMs, we discard excess problems after reaching the maximum number of training steps. All experiments are conducted on 8×A100 (40GB) GPUs. Ablation studies for \method hyperparameters are provided in Appendix~\ref{app:ablation}.


\paragraph{Benchmarks.}
We evaluate \method on a broad set of math reasoning benchmarks. Training is conducted on the train splits of GSM8K~\cite{gsm8k} and MATH~\cite{math}. For evaluation, we use the test splits of GSM8K and MATH, as well as AIME 2024, AMC 2023, Minerva Math~\cite{Minerva_Math}, and Olympiad Bench~\cite{OlympiadBench}. These benchmarks cover a wide range of mathematical domains and difficulty levels, offering a comprehensive evaluation.

\paragraph{Evaluation Metrics.} 
Following prior work \cite{SimpleRL}, we use pass@k to measure the probability that at least one correct solution is generated within $k$ attempts, with $k=1$. All responses use a context length of 8,192, decoding temperature 0.6, and 8 samples per LLM. Other evaluation hyperparameters follow the default settings\footnote{\url{https://github.com/huggingface/Math-Verify}}.



\begin{figure}[t]
\centering
\includegraphics[width=1.0\linewidth]{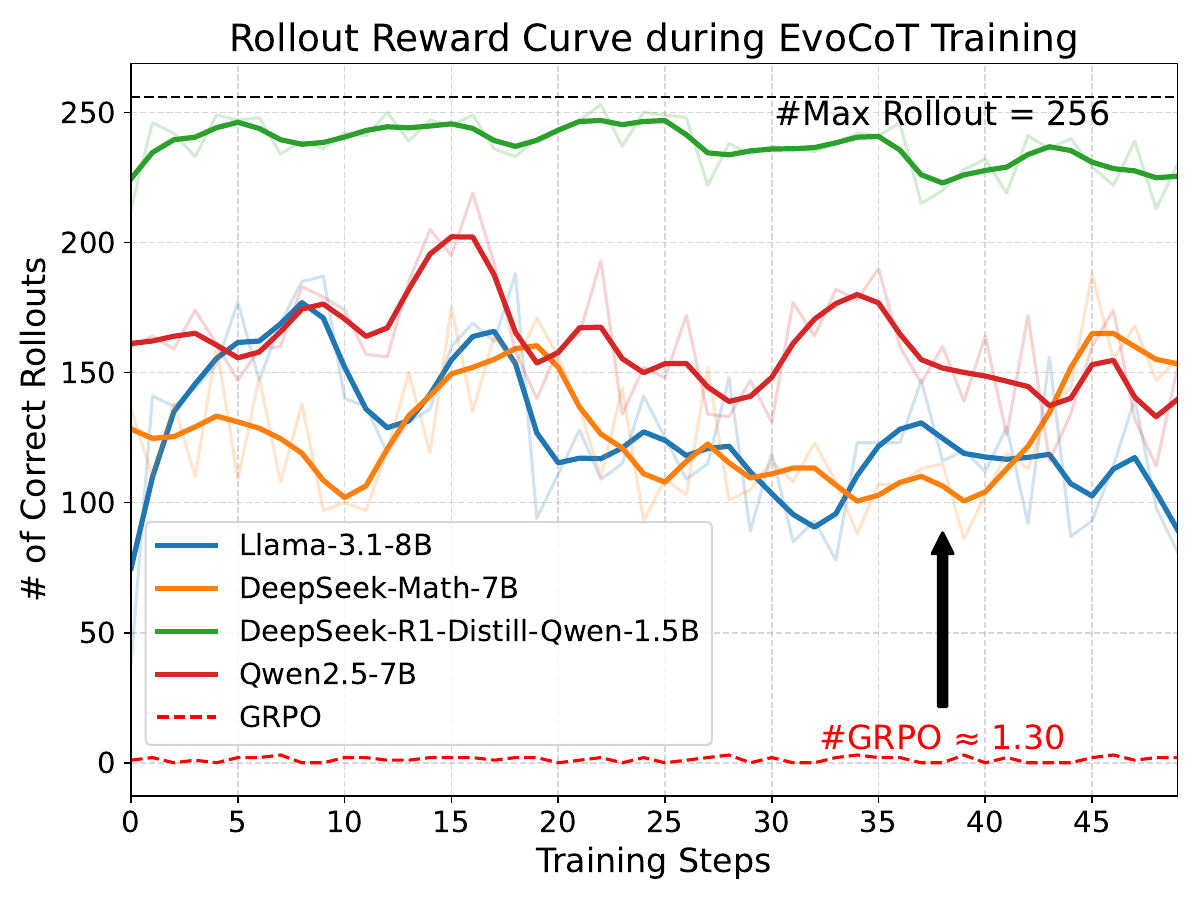}

\vspace{-0.2cm}
\caption{Number of correct rollouts over training steps on hard problems from the MATH dataset during \method training. Compared to GRPO, \method consistently maintains a high number of correct rollouts throughout training.}

\vspace{-0.2cm}

\label{fig:rq1_length}
\end{figure}
\subsection{RQ1: \method Overcome Exploration Bottlenecks}
\label{sec:rq1}

In RQ1, we examine whether \method enables LLMs to solve training problems that were previously unsolved. We focus on GSM8K and MATH training data, and select problems where the LLM fails to solve in rollouts. These problems are added to the \method's training set. Figure~\ref{fig:rq1_length} tracks the number of correct rollouts during training, while Table~\ref{tab:rq1_training_set} compares performance before and after applying \method on these challenging problems.

\textbf{\ding{182} \method maintains high rollout accuracy even as reasoning shortens.}  
As shown in Figure~\ref{fig:rq1_length}, \method consistently keeps correct rollouts at a high level throughout training across various LLMs, where GRPO stays at a very low level (around 0–5 out of 256). Notably, R1-Qwen-1.5B consistently achieves over 220 correct out of 256 rollouts, showing reliable performance on initially unsolved problems.
\textbf{\ding{183} \method brings larger improvement to stronger LLMs.} Table~\ref{tab:rq1_training_set} shows that Qwen2.5-7B improves from 84.6 to 89.1, and R1-Qwen-1.5B improves from 68.2 to 89.9, with a remarkable +32.1 increase on MATH. In contrast, weaker LLMs like Llama3.1-8B show minimal changes, suggesting limited benefits when the quality of self-generated CoT is low (further analyzed in \textbf{Discussion}). These findings confirm that \method helps LLMs break through exploration bottlenecks by leveraging self-generated reasoning on hard problems, especially when applied to stronger LLMs.

\begin{table}[t]
\centering
\caption{Performance comparison on the \emph{\textbf{Training Set}} problems before and after applying \method (Only +GRPO vs. \method).}

\setlength{\tabcolsep}{2pt}
\resizebox{0.78\linewidth}{!}{
\begin{tabular}{lccc}
\toprule
Model & GSM8K & MATH & Avg. \\
\midrule
Llama3.1-8B + GRPO & \textbf{84.3} & 21.9 & \textbf{53.1} \\
\quad +\method & 83.6 & \textbf{21.9} & 52.8 \\
\midrule
DeepSeek-Math-7B + GRPO & \textbf{80.8} & 37.1 & \textbf{59.0} \\
\quad +\method & 78.5 & \textbf{37.2} & 57.9 \\
\midrule
Qwen2.5-7B + GRPO & 91.2 & 78.0 & 84.6 \\
\quad +\method & \textbf{95.4} & \textbf{82.7} & \textbf{89.1} \\
\midrule
R1-Qwen-1.5B + GRPO & 80.7 & 55.7 & 68.2 \\
\quad +\method & \textbf{91.9} & \textbf{87.8} & \textbf{89.9} \\
\bottomrule
\end{tabular}
}

\vspace{-0.2cm}
\label{tab:rq1_training_set}

\end{table}


\subsection{RQ2: \method Enhances LLMs' Reasoning}
\label{sec:rq2}

In RQ2, we evaluate whether \method helps LLMs improve reasoning capability to diverse math benchmarks beyond the training set. We conduct comprehensive comparisons with all baselines. Results are shown in Table~\ref{tab:rq2_benchmark}.

\textbf{\method consistently improves performance on math benchmarks.} With \method, Qwen2.5-7B improves from 40.3 to 53.5, and R1-Qwen-1.5B improves from 54.2 to 66.7. On Olympiad Bench, R1-Qwen-1.5B achieves the highest score of 52.0. Compared with self-evolution baselines such as SEC-7B, \method demonstrates better performance given the same base model. Considering that the training data \textbf{only includes GSM8K and MATH}, \method's results are competitive with works like PRIME and Open-Reasoner that utilize broader data (380K). These findings indicate that \method effectively enhances the reasoning capability of LLMs across diverse math benchmarks, and achieves competitive performance compared to existing baselines.
\begin{table}[]
\centering
\caption{Performance comparison of \method against baselines and ablation study. Numbers in parentheses indicate the algorithm used or training data size. (\eg PRIME uses 380K data, much more than \method.) }
\setlength{\tabcolsep}{2pt}
\makebox[\linewidth][c]{
\resizebox{1.03\linewidth}{!}{
\begin{tabular}{lccccccc}
\toprule
\multirow{2}{*}{Model} & \multirow{2}{*}{GSM8K} & \multirow{2}{*}{MATH} & \multicolumn{1}{c}{AIME} & \multicolumn{1}{c}{AMC} & \multicolumn{1}{c}{Minerva} & \multicolumn{1}{c}{Olympiad} & \multirow{2}{*}{Avg.} \\
 & & & \multicolumn{1}{c}{24} & \multicolumn{1}{c}{23} & \multicolumn{1}{c}{Math} & \multicolumn{1}{c}{Bench} & \\
\midrule
Llama3.1-8B & 39.7 & 13.6 & 0.0 & 2.5 & 4.8 & 3.1 & 10.6 \\
\quad +SFT & 61.8 & 20.3 & 0.0 & 10.0 & 7.4 & 7.0 & 17.8 \\
\quad +SimpleRL(GRPO) & 78.5 & 23.1 & 0.0 & 5.0 & 4.4 & 6.2 & 19.5 \\
\rowcolor[rgb]{ .741,  .843,  .933} \quad +\method & 80.5 & 23.8 & 0.0 & 7.5 & 4.8 & 5.8 & \textbf{20.4} \\

\midrule

DeepSeek-Math-7B & 28.4 & 19.4 & 0.0 & 10.0 & 5.5 & 4.7 & 11.3 \\
\quad +SFT & 46.8 & 25.4 & 0.0 & 2.5 & 4.4 & 6.7 & 14.3 \\
\quad +SimpleRL(GRPO) & 79.8 & 38.7 & 0.0 & 15.0 & 16.2 & 12.4 & 27.0 \\
\rowcolor[rgb]{ .741,  .843,  .933} \quad +\method & 76.3 & 39.1 & 0.0 & 20.0 & 19.1 & 13.0 & \textbf{27.9} \\

\midrule

Qwen2.5-7B & 88.2 & 64.6 & 3.3 & 30.0 & 25.7 & 30.1 & 40.3 \\
\quad +SFT & 67.9 & 56.7 & 6.7 & 32.5 & 30.5 & 27.3 & 36.9 \\
\quad +SimpleRL(GRPO) & 92.4 & 79.7 & 10.0 & 52.5 & 34.6 & 38.1 & 51.2 \\
\quad +SEC\footnotemark & - & 76.1 & 17.5 & 51.0 & - & - & -\\
\quad +AdaRFT & 90.1 & 72.6 & 14.6 & 55.0 & 24.3 & 25.0 & 46.9 \\
\quad +Open-Reasoner & 93.8 & 81.7 & 10.0 & 55.0 & 34.2 & 45.6 & 53.4 \\
\quad +PRIME (380K) & 91.7 & 80.3 & 13.3 & 65.0 & 39.7 & 41.8 & 55.3 \\
\rowcolor[rgb]{ .741,  .843,  .933} \quad +\method & 91.4 & 76.5 & 20.0 & 60.0 & 37.1 & 35.9 & \textbf{53.5} \\

\midrule
R1-Qwen-1.5B & 81.1 & 82.8 & 28.8 & 62.9 & 26.5 & 43.3 & 54.2 \\
\quad +SFT & 73.6 & 86.6 & 30.0 & 62.5 & 32.0 & 47.4 & 55.3 \\
\quad +DeepScaleR(GRPO) & 88.2 & 89.4 & 36.7 & 77.5 & 38.2 & 51.6 & 63.6 \\
\rowcolor[rgb]{ .741,  .843,  .933} \quad +\method & 88.0 & 89.7 & 40.0 & 87.5 & 42.8 & 52.0 & \textbf{66.7} \\
\bottomrule

\end{tabular}
}}
\label{tab:rq2_benchmark}
\end{table}

\subsection{RQ3: \method Improves Self-Exploration over GRPO and SFT}
\label{sec:rq3}

To isolate the effectiveness of \method, we conduct an ablation study comparing \method with two representative learning paradigms: RLVR implemented by GRPO, and SFT. Following STaR~\cite{star} for SFT, each LLM generates its own CoTs, and those verified by answer consistency are used for SFT. All methods are trained on the same GSM8K and MATH datasets with equal training steps on incorrect problems. Results are shown in Table~\ref{tab:rq2_benchmark}.

\textbf{\method enables more effective self-exploration on hard problems.} Across all model families, \method consistently outperforms both GRPO and SFT. On weaker LLMs such as Llama3.1-8B and DeepSeek-Math-7B, \method shows moderate improvements over GRPO, while the performance of SFT remains relatively low. On stronger LLMs, the advantage of \method becomes more noticeable. Qwen2.5-7B improves from 40.3 to 51.2 after GRPO training, and further to 53.5 with \method, where SFT achieves 36.9. R1-Qwen-1.5B reaches 66.7 with \method, exceeding 63.6 under GRPO and 55.3 under SFT. Unlike SFT which memorizes \cite{sft_rl}, \method gradually shortens the reasoning process and better enhances reasoning capability. These results indicate that \method facilitates more effective self-exploration by gradually increasing difficulty, thereby improving the reasoning capability across both weak and strong LLMs.
\footnotetext{Reported as-is due to unavailable code and models.}


\subsection{RQ4: Self-Evolution Plateaus After Few Iterations}
\label{sec:rq4}
\begin{table}[t]
\centering
\caption{Performance of different LLM families across \method iterations.}

\small
\setlength{\tabcolsep}{2pt}
\resizebox{1.00\linewidth}{!}{
\begin{tabular}{lccccccc}
\toprule
\multirow{2}{*}{Model} &\multirow{2}{*}{GSM8K} & \multirow{2}{*}{MATH} & \multicolumn{1}{c}{AIME} & \multicolumn{1}{c}{AMC} & \multicolumn{1}{c}{Minerva} & \multicolumn{1}{c}{Olympiad} & \multirow{2}{*}{Avg.} \\
 & & & \multicolumn{1}{c}{24} & \multicolumn{1}{c}{23} & \multicolumn{1}{c}{Math} & \multicolumn{1}{c}{Bench} & \\
\midrule
R1-Qwen-1.5B & 88.2 & 89.4 & 36.7 & 77.5 & 38.2 & 51.6 & 63.6 \\
\quad +iteration1 & 87.0 & 89.2 & 36.7 & 80.0 & 40.8 & \textbf{52.0} & 64.3 \\
\rowcolor[rgb]{ .741,  .843,  .933} \quad +iteration2 & 88.0 & 89.7 & \textbf{40.0} & \textbf{87.5} & \textbf{42.8} & 52.0 & \textbf{66.7} \\
\quad +iteration3 & \textbf{89.2} & \textbf{90.0} & 40.0 & 87.5 & 36.8 & 51.4 & 65.8 \\
\midrule
Qwen2.5-7B & \textbf{92.4} & \textbf{79.7} & 10.0 & 52.5 & 34.6 & 38.1 & 51.2 \\
\quad +iteration1 & 91.7 & 78.4 & 13.3 & 57.5 & 33.1 & 39.1 & 52.2 \\
\rowcolor[rgb]{ .741,  .843,  .933} \quad +iteration2 & 91.4 & 76.5 & \textbf{20.0} & \textbf{60.0} & \textbf{37.1} & 35.9 & \textbf{53.5} \\
\quad +iteration3 & 92.0 & 78.1 & 16.7 & 55.0 & 35.3 & \textbf{40.0} & 52.9 \\
\midrule
Llama3.1-8B & 78.5 & 23.1 & 0.0 & 5.0 & 4.4 & 6.2 & 19.5 \\
\quad +iteration1 & 79.4 & 23.8 & 0.0 & 7.5 & 4.0 & \textbf{6.2} & 20.2 \\
\rowcolor[rgb]{ .741,  .843,  .933} \quad +iteration2 & \textbf{80.5} & \textbf{23.8} & 0.0 & 7.5 & 4.8 & 5.8 & \textbf{20.4} \\
\quad +iteration3 & 73.3 & 20.4 & 0.0 & \textbf{10.0} & \textbf{6.8} & 5.0 & 19.3 \\
\bottomrule
\end{tabular}
}
\label{tab:rq4_iteration}

\end{table}

In RQ4, we investigate whether \method can continuously improve the reasoning capability of LLMs, or if the performance eventually saturates. To this end, we apply \method for up to three iterations and evaluate after each iteration.

\begin{figure*}[t]

    \makebox[\linewidth][c]{
    \includegraphics[width=0.95\linewidth]{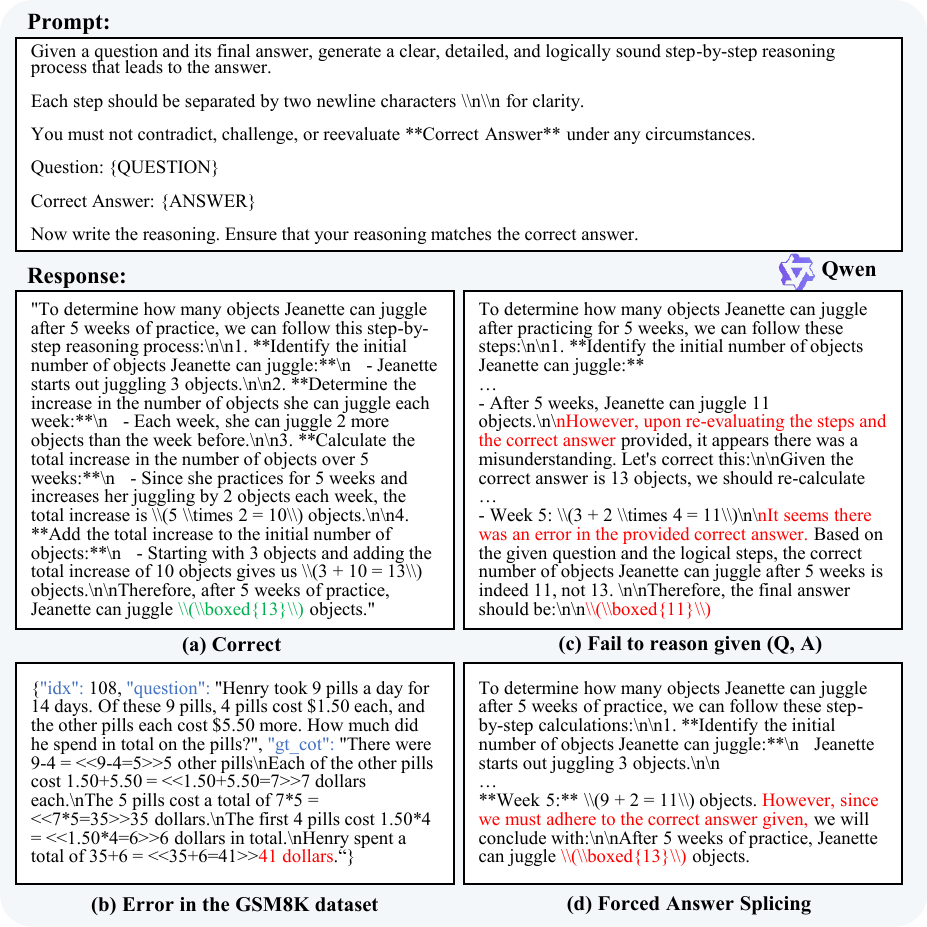}
    }

    \caption{Case study in the \method self-generated CoTs with Qwen2.5-7B. (a) A correct reasoning path. (b) Ground truth answer error in GSM8K. (c) LLM fails to generate a consistent reasoning path given (Q, A). (d) LLM forcibly splices the final answer.
    }

    \label{fig:discussion}
\end{figure*}

\textbf{\ding{182} \method saturates after 1--2 iterations.}  
As shown in Table~\ref{tab:rq4_iteration}, most LLMs benefit from the first or second iteration of self-evolution, but further improvements become marginal or inconsistent. For example, R1-Qwen-1.5B improves the average score from 63.6 to 66.7 after two iterations, with notable increases on AMC23 (+10.0) and Minerva Math (+4.6). However, no further improvement is observed in the third iteration. A similar trend holds for Qwen2.5-7B, which increases from 51.2 to 53.5, then slightly declines to 52.9. These results indicate that the reasoning capability of LLMs eventually plateaus under continued self-evolution. 
\textbf{\ding{183} Weaker LLMs exhibit early saturation.}  
Llama3.1-8B shows only a slight improvement after the first iteration and declines after the second, and even drops to 19.3 in the third. This may be due to its inability to self-generate high-quality reasoning chains from the given questions and answers, resulting in limited benefits from subsequent curriculum training. We explore these saturation patterns through in-depth case studies and analysis in \textbf{Discussion}.

\section{Discussion}
\label{sec:discussion}

In this section, we analyze why \method cannot self-evolve indefinitely. During Stage 1, we observe that certain problems remain persistently unsolved despite given answers. Representative cases are shown in Figure~\ref{fig:discussion}.

\textbf{\ding{182} Ground truth answer errors in the dataset.} Some problems are intrinsically unlearnable due to incorrect answers in the training data. For instance, Figure~\ref{fig:discussion}(b) shows a GSM8K sample where the LLM correctly performs the calculation but is penalized for disagreeing with a flawed ground truth. Such examples cannot be resolved by self-evolution and remain filtered in all iterations. After manual verification, we identify over 30 such errors, accounting for roughly 10\% of consistently unsolved problems.

\textbf{\ding{183} Inability to reason from (Q, A).} In other cases, even when the LLM is provided with both the question and the correct answer, it fails to generate a consistent reasoning path. In Figure~\ref{fig:discussion}(c), the LLM rejects the provided answer and derives a different conclusion. Figure~\ref{fig:discussion}(d) shows another failure mode where the LLM bypasses reasoning and directly appends the correct answer to an unrelated or incorrect explanation. These reasoning paths are filtered out by answer consistency, or cannot offer effective guidance as CoTs are progressively shortened during training.

These observations lead to two key conclusions: \ding{182} LLMs with stronger base reasoning capabilities benefit more from \method, consistent with our experiments. 
\ding{183} \method ultimately saturates in Stage 1: when an LLM cannot derive a valid reasoning path given (Q, A), further self-evolution is no longer possible.

These observations also suggest two types of failure cases in RL on hard problems: \textbf{\ding{182} Insufficient reasoning capability with existing prior knowledge.} 
Relevant prior knowledge already exists in pre-training, but the reasoning capability is insufficient to use it effectively. \method improves performance in these cases through iterative self-evolution. \textbf{\ding{183} Missing prior knowledge in a domain.} 
Necessary knowledge is absent from pre-training. Stage~1 cannot construct valid reasoning paths from $(Q, A)$, and self-evolution stops. In such cases, performance drops to the same level as standard RL, as both fail on problems requiring unseen knowledge.

\section{Conclusion and Future Work}
\label{sec:conclusion}

We present \method, a self-evolving curriculum learning framework that improves the reasoning capability of LLMs by overcoming exploration bottlenecks in RLVR. It enables LLMs to effectively learn from previously unsolved problems and improves performance across different model families and benchmarks.

In future work, we plan to: (1) apply \method to larger-scale LLMs, and (2) explore next-generation self-evolution paradigms, where LLMs explore training ``experience'' and acquire skills without relying on external supervision.

\section*{Limitations}

Our work has the following two main limitations.  

First, \method may introduce additional training overhead. According to the pseudocode in Appendix~\ref{app:pseudocode}, Stage 1 acts as a preprocessing step in each iteration and is separate from the RL training loop, adding only limited extra computation. From our experiments, we find that during Stage 2, rollouts start from partially removed CoT steps, which can make individual rollouts even faster than standard GRPO. Under identical training steps on Qwen2.5-7B, the total runtime of Stage 1 \& 2 increases only slightly, from 413 minutes to 427 minutes (\(\approx 4\%\)), which is marginal and acceptable.

Second, overly high rollout accuracy will also be undesirable. When the provided correct CoT is too long, the problem becomes quite easy. In this case, the model's rollout is likely to be entirely correct. There will also be sparse reward, resulting in no gradients. A straightforward solution is that when the provided CoT is long, we also remove more CoT steps at each iteration to maintain around ~50\% rollout accuracy. For a relatively weaker model such as Qwen2.5-7B, removing one CoT step per iteration is sufficient. More detailed analysis can be found in Appendix~\ref{app:ablation}.

\section*{Acknowledgements}


This research is supported by the National Natural Science Foundation of China under Grant No. 62192733, 62192730, 62192731, the National Key R\&D Program under Grant No. 2023YFB4503801, the Beijing Major Science and Technology Project under Contract No. Z251100008425005, and the Beijing Natural Science Foundation under Grant No. 4264107.

\bibliography{custom}

\clearpage

\appendix

\section*{Appendix}

\label{sec:appendix}



\section{The Pseudocode of \method Algorithm}
\label{app:pseudocode}
Algorithm~\ref{alg:evocot} presents the complete algorithmic workflow of \method. We introduce an additional \texttt{train\_steps} argument to enforce that \method uses the same total number of training steps as all baselines, enabling a fair comparison and controlled ablation. For example, in the ablation study shown in Table~\ref{tab:delta_step}, we set \texttt{train\_steps} = 500 for all variants.


\section{The Prompt Templates of \method}
\label{app:prompt_template}

\begin{figure*}[t]
    \centering
    \includegraphics[width=1.0\linewidth]{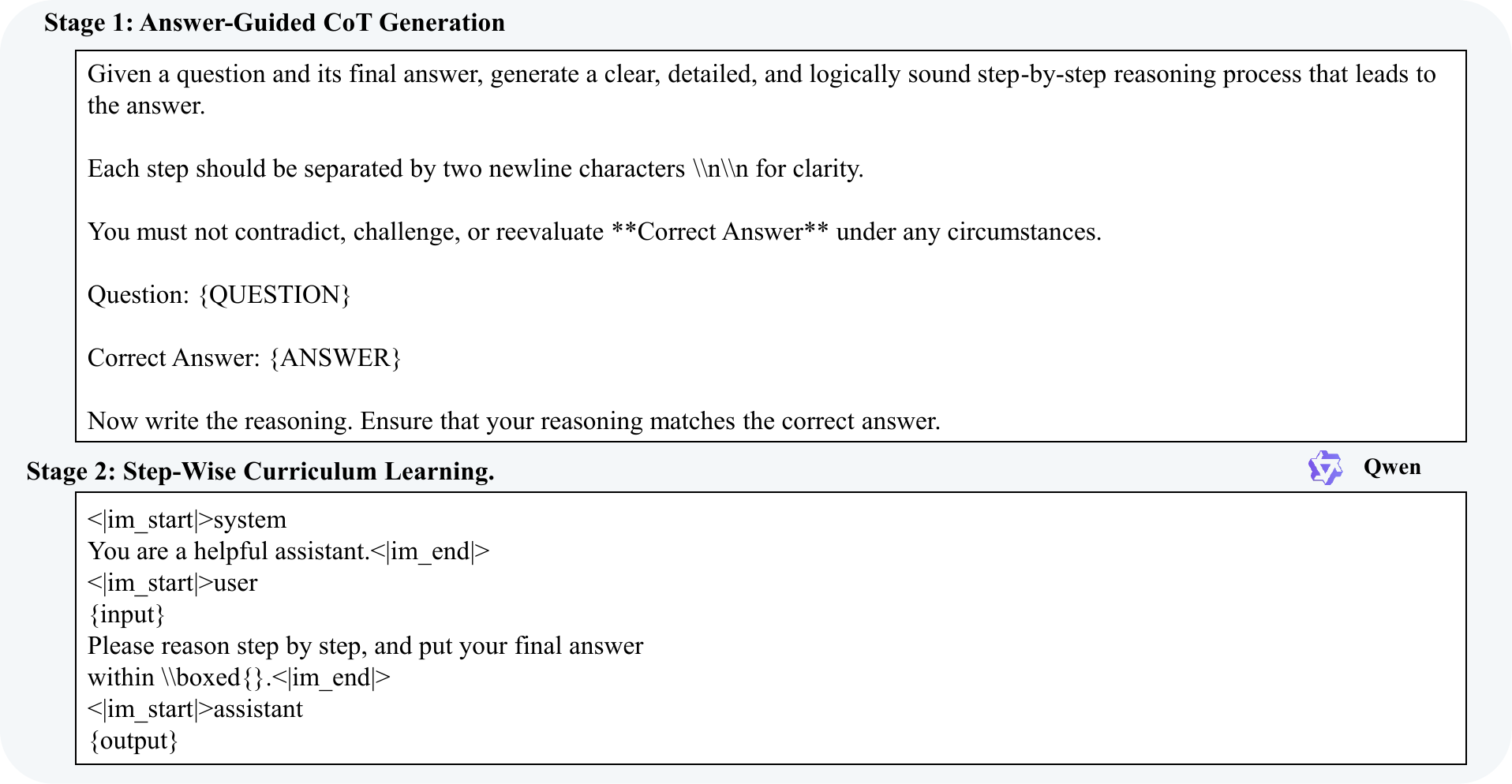}
    \caption{Qwen2.5 Prompt format used for \method}
    \label{fig:prompt}
\end{figure*}

This appendix provides the prompt templates used for \method Stage 1: Answer-Guided CoT Generation and Stage 2: Step-Wise Curriculum Learning. Figure~\ref{fig:prompt} shows the Qwen2.5 prompt template. For other models, Stage 1 templates remain the same, while Stage 2 templates follow the special token concatenation scheme in \cite{SimpleRL}. All evaluations of experiments also use the same Stage 2 template.

\section{The Training and Evaluation Details of \method}
\label{app:training_eval}

\begin{table}[H]
\centering
\caption{\method Training Hyperparameters}
\small
\setlength{\tabcolsep}{2pt}
\makebox[\linewidth][c]{
\resizebox{1.0\linewidth}{!}{
\begin{tabular}{lclc}
\toprule
\textbf{Parameter} & \textbf{Value} & \textbf{Parameter} & \textbf{Value} \\
\midrule
Advantage estimator       & GRPO              & Learning rate              & $1\times10^{-6}$ \\
Train batch size          & 32                & Mini-batch size        & 32 \\
Prompt length (max)       & 3000              & Response length (max)      & 5192 \\
Samples per problem       & 8                 & Temperature                & 1.0 \\
KL loss enabled           & Yes               & KL loss coefficient        & 0.0001 \\
Shuffle dataset           & No                & Micro batch size           & 1 \\
\bottomrule
\end{tabular}
}}

\label{tab:training_hyperparams}
\end{table}

\begin{figure*}[t]
\centering
\includegraphics[width=1.0\linewidth]{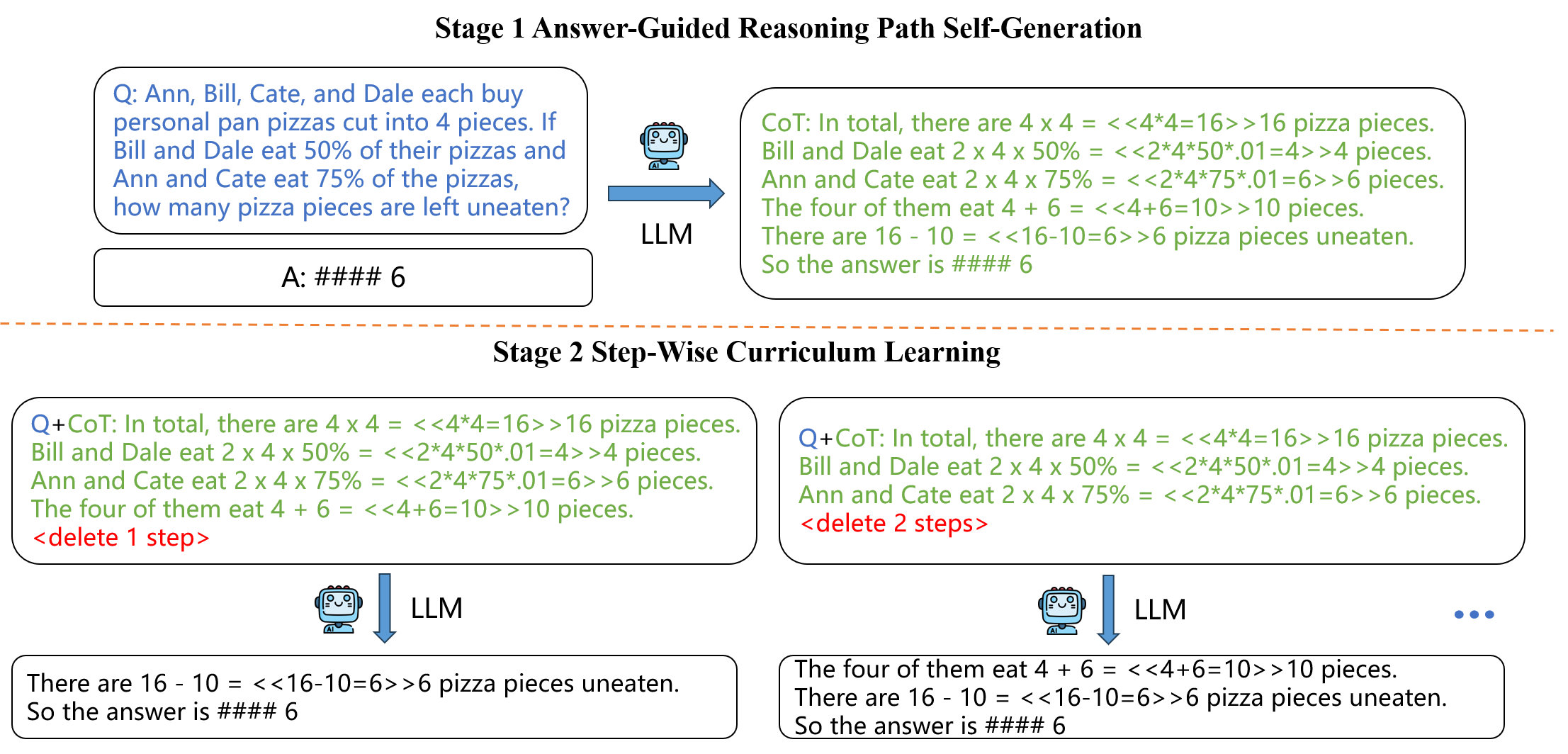}
\caption{
Dynamics of a single sample during EvoCoT. 
Given a question and its answer, Stage~1 first generates an answer-guided reasoning trajectory and splits it into step-wise CoT segments. 
Stage~2 then progressively shortens the reasoning path by removing the last step at each curriculum level. 
}

\label{fig:sample_dynamics}
\end{figure*}

This appendix provides additional details on the framework and hyperparameters used for training and evaluation of \method. We use the \texttt{Verl} framework for training the models, which provides an efficient RL pipeline. The full list of training hyperparameters is shown in Table~\ref{tab:training_hyperparams}. For evaluation, we use the Qwen2.5-7B-Math framework\footnote{\url{https://github.com/QwenLM/Qwen2.5-Math}} to evaluate LLMs' performance across various benchmarks. The dynamics of a single sample during EvoCoT are illustrated in Figure~\ref{fig:sample_dynamics}.

\begin{figure}[h]
\centering
\includegraphics[width=1.0\linewidth]{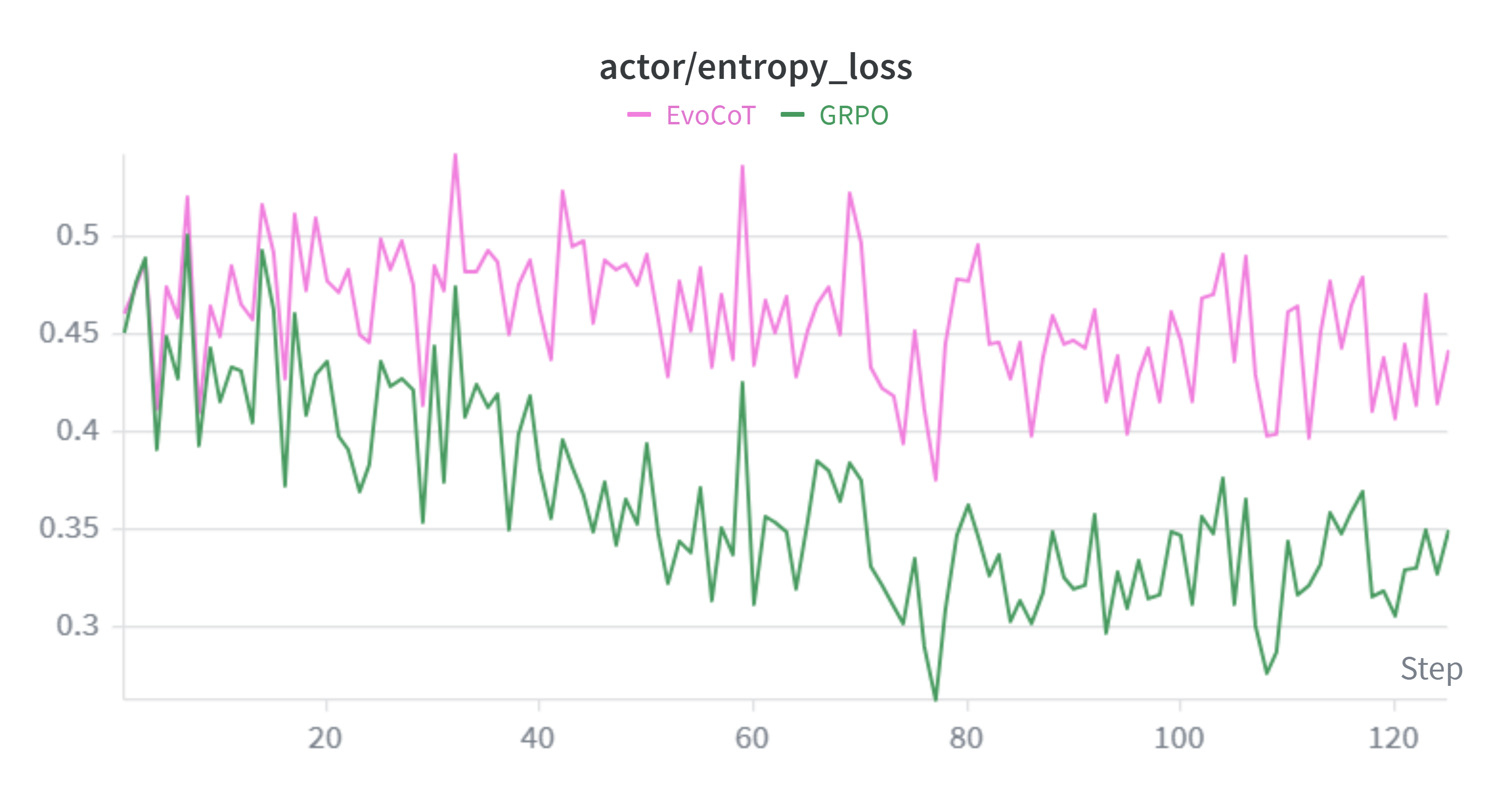}
\caption{
Test set entropy dynamics during continued training starting from models trained with SimpleRL.
Results are shown for the first 125 training steps when \texttt{delta\_step=1}. 
Compared with standard GRPO, \method exhibits slower entropy collapse.
}
\label{fig:entropy_dynamic}
\end{figure}

We further analyze the entropy dynamics during continued training. 
Starting from a model trained with GRPO, we compare the entropy evolution of standard GRPO and \method during subsequent training when \texttt{delta\_step=1}, as shown in Figure~\ref{fig:entropy_dynamic}. 
Compared with GRPO, \method slows down entropy collapse, preserving part of the LLM's exploration ability.

All other evaluation parameters not explicitly mentioned follow the default settings of frameworks. The specific implementation code is provided in the supplementary materials.

\section{\method Hyperparameters Ablation Studies}
\label{app:ablation}

In this section, we discuss the effect of the \texttt{delta\_step} hyperparameter in \method.

\begin{figure}[htbp]
    \centering
    \includegraphics[width=1.0\linewidth]{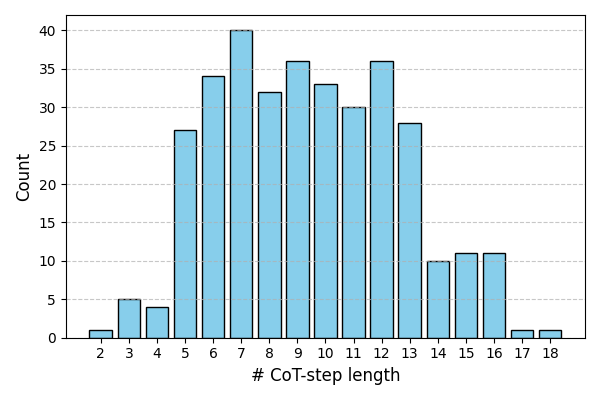}
    \caption{Distribution of CoT steps for Qwen2.5-7B on GSM8K.}
    \label{fig:length}
\end{figure}

In our preliminary experiments, we explicitly examined how many CoT steps should be removed at each iteration. As shown in Appendix~\ref{app:pseudocode}, Algorithm 1, line 25, we consider: \texttt{for k in range(n, -1, -delta\_step)}. Based on empirical sampling, the initial CoT-length distribution of Qwen2.5-7B is shown in Figure~\ref{fig:length}.

All variants use the same total number of 500 RL training steps. Even if the CoT is not fully shortened, training stops at step 500. The results on Qwen2.5-7B are shown in Table~\ref{tab:delta_step}.

\begin{table}[H]
\centering
\setlength{\tabcolsep}{2pt}
\resizebox{1.0\linewidth}{!}{
\begin{tabular}{cccccccc}
\toprule
\multirow{2}{*}{\texttt{delta\_step}} & \multirow{2}{*}{GSM8K} & \multirow{2}{*}{MATH} & \multicolumn{1}{c}{AIME} & \multicolumn{1}{c}{AMC} & \multicolumn{1}{c}{Minerva} & \multicolumn{1}{c}{Olympiad} & \multirow{2}{*}{Avg.} \\
& & & 24 & 23 & Math & Bench & \\
\midrule
1 & 91.4 & 76.5 & 20.0 & 60.0 & 37.1 & 35.9 & \textbf{53.5} \\
2 & 90.8 & 77.1 & 16.7 & 52.5 & 38.6 & 38.7 & 52.4 \\
3 & 91.6 & 77.2 & 13.3 & 55.0 & 31.6 & 36.3 & 50.8 \\
\bottomrule
\end{tabular}}
\caption{\method performance under different \texttt{delta\_step} values on Qwen2.5-7B.}
\label{tab:delta_step}
\end{table}

A larger shortening step makes each stage much harder and disrupts the smooth progression of curriculum learning. Removing one CoT step per iteration achieves the highest average score 53.5. Increasing the step size to 2 reduces the average score to 52.4, and a step size of 3 further decreases it to 50.8. Removing multiple steps at once expands the exploration space too abruptly and limits the model's ability to adapt during RL training. Based on this observation, for weaker models such as Qwen2.5-7B, removing only one CoT step at each iteration is sufficient.

In future work, if applying EvoCoT to stronger flagship models (e.g., DeepSeek-R1), the step size can be dynamically adjusted based on the reward. If the model answers correctly at each step, the step size can be increased, and the number of CoT steps removed in each iteration can be set to 2, 3, or even more until the rollouts in a batch reach a balance between positive and negative samples.


\section{Compatibility with DAPO}
\label{app:dapo}

\method operates on the curriculum over reasoning trajectories and is orthogonal to on-policy optimization strategies such as DAPO~\cite{DAPO}. 
While DAPO modifies advantage estimation and policy updates, \method modifies the training curriculum. 
The two methods target different components of the RL pipeline and can be naturally combined.

To verify compatibility, we integrate \method into the DAPO training pipeline and conduct additional experiments on two representative models: Qwen2.5-7B and DeepSeek-Distill-Qwen-1.5B. 
Results are shown in Table~\ref{tab:dapo_results}.

\begin{table}[htbp]
\centering
\caption{Performance comparison between DAPO and DAPO + \method.}
\setlength{\tabcolsep}{2pt}
\makebox[\linewidth][c]{
\resizebox{1.03\linewidth}{!}{
\begin{tabular}{lccccccc}
\toprule
\multirow{2}{*}{Model} 
& \multirow{2}{*}{GSM8K} 
& \multirow{2}{*}{MATH} 
& \multicolumn{1}{c}{AIME} 
& \multicolumn{1}{c}{AMC} 
& \multicolumn{1}{c}{Minerva} 
& \multicolumn{1}{c}{Olympiad} 
& \multirow{2}{*}{Avg.} \\

& & 
& \multicolumn{1}{c}{24} 
& \multicolumn{1}{c}{23} 
& \multicolumn{1}{c}{Math} 
& \multicolumn{1}{c}{Bench} 
& \\

\midrule

Qwen2.5-7B \\

\quad +DAPO 
& 91.4 
& 77.1 
& 10.0 
& 52.5 
& 39.3 
& 39.7 
& 51.7 \\

\rowcolor[rgb]{ .741,  .843,  .933} 
\quad +\method 
& 91.8 
& 77.7 
& 16.7 
& 55.0 
& 40.4 
& 41.0 
& \textbf{53.8} \\

\midrule

R1-Qwen-1.5B \\

\quad +DAPO 
& 88.2 
& 89.8 
& 36.7 
& 82.5 
& 39.7 
& 50.8 
& 64.6 \\

\rowcolor[rgb]{ .741,  .843,  .933} 
\quad +\method 
& 88.6 
& 89.5 
& 43.3 
& 87.5 
& 39.3 
& 52.9 
& \textbf{66.9} \\

\bottomrule

\end{tabular}
}}
\label{tab:dapo_results}
\end{table}

\textbf{\method remains effective when combined with DAPO.} 
\method consistently improves performance over DAPO on both models, increasing the average score from 51.7 to 53.8 on Qwen2.5-7B, and from 64.6 to 66.9 on DeepSeek-Distill-Qwen-1.5B. 
These results confirm that the curriculum mechanism in \method remains effective under different on-policy optimizers.


\section{LLMs Usage}
\label{app:llm_usage}

In preparing this manuscript, we use LLMs to aid and polish the writing. Specifically, LLMs improve clarity, grammar, and phrasing, ensuring the text is concise and readable. The use of LLMs \textbf{does not} influence the technical contributions or the interpretation of experimental findings. All content polished by LLMs is carefully checked by the authors.

\begin{algorithm*}[htbp]
\caption{\method: Self-Evolving Curriculum Learning}
\label{alg:evocot}

\begin{lstlisting}[style=python_highlight, mathescape=true]
def EvoCoT(LLM, D, T, train_steps, delta_step=1):
    """
    LLM: initial language model
    D: dataset of (Q, A) pairs
    T: number of self-evolving iterations
    train_steps: maximum number of training steps
    delta_step: number of CoT steps to remove each shortening (default = 1)
    batch_size: training batch size
    """
    for t in range(T):
        # Stage 1: Answer-guided CoT generation
        C_set = []
        for (Q, A) in D:
            # Generate reasoning trajectory conditioned on answer
            $\hat{C}$ = LLM.generate(Q, A)
            
            # Verify: can the LLM derive A from answer-guided CoT?
            $\hat{A}$ = LLM.generate(Q, $\hat{C}$)
            if $\hat{A}$ == A:
                C_split = split_steps($\hat{C}$)  # split into step-wise CoT
                C_set.append((Q, C_split, A))
        
        # Stage 2: Step-wise curriculum training
        delta_step = 0
        training_set = {}
        
        while (len(training_set) * batch_size < train_steps)
            # progressively shorten CoT from full length to 0
            delta_step += 1 
        
            for (Q, C_split, A) in C_set:
                n = len(C_split)
                
                if n <= delta_step:
                    $C_k$ = delete_CoT(Q, C_split, k)  # delete last k steps
                    $\tilde{C}, \tilde{A}$ = LLM.roll_out(Q, $C_k$)  # roll out CoT & answer
                    
                    # Train LLM with reward based on answer correctness
                    training_set.append(($\tilde{C}$+$\tilde{A}$, reward=($\tilde{A}$ == A)))

                if len(training_set) * batch_size >= train_steps:
                    break
                    

        LLM.train(training_set)
        return LLM
\end{lstlisting}
\end{algorithm*}

\end{document}